\documentclass{article}

\usepackage[final]{neurips_2025}
\usepackage{graphicx}    
\usepackage{svg}         
\usepackage{amssymb}

\usepackage[utf8]{inputenc} 
\usepackage[T1]{fontenc}    
\usepackage{hyperref}       
\usepackage{url}            
\usepackage{booktabs}       
\usepackage{amsfonts}       
\usepackage{nicefrac}       
\usepackage{microtype}      
\usepackage{xcolor}         
\usepackage{xspace}
\usepackage{amsmath} 
\usepackage{adjustbox}
\usepackage{multicol}
\usepackage{multirow}
\usepackage{colortbl}

\newcommand{\modelname}{\textsc{OmniDPO}\xspace}

\title{\modelname: A Preference Optimization Framework to Address Omni-Modal Hallucination}

\author{
\textbf{Junzhe Chen}$^{1,2*}$ \And 
\textbf{Tianshu Zhang}$^{1*}$ \And 
\textbf{Shiyu Huang}$^{3}$ \And 
\textbf{Yuwei Niu}$^{4}$ \AND
\textbf{Chao Sun}$^{1}$ \And
\textbf{Rongzhou Zhang}$^{1}$ \And 
\textbf{Guanyu Zhou}$^{2}$ \And
\textbf{Lijie Wen}$^{1\dagger}$ \And 
\textbf{Xuming Hu}$^{2\dagger}$ \\
$^1$Tsinghua University, 
$^2$The Hong Kong University of Science and Technology (Guangzhou), \\
$^3$OpenRL, 
$^4$Chongqing University, \\
\texttt{chenjz24@mails.tsinghua.edu.cn} \\
\texttt{wenlj@tsinghua.edu.cn, xuminghu97@gmail.com}
}

\begin{document}

\maketitle
\begin{abstract}
Recently, Omni-modal large language models (OLLMs) have sparked a new wave of research, achieving impressive results in tasks such as audio-video understanding and real-time environment perception. However, hallucination issues still persist. Similar to the bimodal setting, the priors from the text modality tend to dominate, leading OLLMs to rely more heavily on textual cues while neglecting visual and audio information. In addition, fully multimodal scenarios introduce new challenges. Most existing models align visual or auditory modalities with text independently during training, while ignoring the intrinsic correlations between video and its corresponding audio. This oversight results in hallucinations when reasoning requires interpreting hidden audio cues embedded in video content. To address these challenges, we propose \modelname, a preference-alignment framework designed to mitigate hallucinations in OLLMs. Specifically, \modelname incorporates two strategies: (1) constructing text-preference sample pairs to enhance the model’s understanding of audio-video interactions; and (2) constructing multimodal-preference sample pairs to strengthen the model’s attention to visual and auditory information. By tackling both challenges, \modelname effectively improves multimodal grounding and reduces hallucination. Experiments conducted on two OLLMs demonstrate that OmniDPO not only effectively mitigates multimodal hallucinations but also significantly enhances the models' reasoning capabilities across modalities. All code and datasets will be released upon paper acceptance.
\end{abstract}

\graphicspath{{figures/}} 

\section{Introduction}
Omni-modal large language models (OLLMs) have rapidly advanced in their ability to understand and generate content from combined inputs such as images, audio, and text. By leveraging information from multiple modalities, these models can perform complex tasks ranging from video question answering to audio-visual scene description. However, despite these advancements, hallucination remains a critical issue: models often produce outputs that do not accurately reflect the given visual or auditory input \citep{nishimura2024audiohallucinationslargeaudiovideo,sahoo2024comprehensivesurveyhallucinationlarge}. This phenomenon of generating inconsistent or fabricated omni-modal content is broadly referred to as omni-modal hallucination. It poses serious risks in high-stakes applications (e.g., medical video analysis or autonomous driving), where factual precision is paramount. 

Previous studies in the bimodal (text-image) setting have shown that the significantly stronger capability of text encoders compared to visual encoders results in dominant textual priors \citep{bai2025hallucinationmultimodallargelanguage,huang2024visualhallucinationsmultimodallarge}. Consequently, models tend to rely heavily on input text while overlooking visual information—a primary cause of multimodal hallucinations \citep{favero2024multimodalhallucinationcontrolvisual}. Similarly, OLLMs exhibit comparable issues: they are inclined to depend on textual inputs while neglecting other modalities \citep{gao2025exploringhallucinationlargemultimodal}. Furthermore, omni-modal scenarios introduce new challenges. Existing OLLMs typically align vision or audio with text separately during training, while failing to account for the intrinsic interactions between video and its associated audio \citep{guo2025alignedbetterlistenbetter}. This limitation leads to hallucinations, especially in tasks that require reasoning based on subtle audio cues embedded within video content.

Research on mitigating multimodal hallucinations generally falls into two categories. 1) Training-free methods, which do not modify the model’s parameters. Representative examples include contrastive decoding (e.g., VCD \citep{leng2024mitigating}), which mitigates hallucinations by introducing blurred or ambiguous images during decoding to expose and counteract biased language priors; and inference-time intervention (e.g., ICT \citep{chen2024ictimageobjectcrossleveltrusted}), which enhances model reliability by injecting intervention vectors into activation layers during the forward pass. However, these methods may inadvertently eliminate all language priors, including those beneficial for reasoning, which can negatively affect performance on reasoning-intensive tasks. Moreover, decoding-based approaches typically require multiple passes during inference, often resulting in significantly increased latency. 2) Training Approaches, which fine-tune models using either synthetic or manually annotated high-quality data to guide the model toward paying more attention to visual information and thus reducing hallucinations. However, manually annotating such data is often prohibitively expensive, while existing synthetic datasets either focus solely on the text modality or fail to capture the intricate relationships between video and its associated audio in omnimodal scenarios.

\begin{figure}[htbp]
  \centering
  \includegraphics[width=0.9\linewidth]{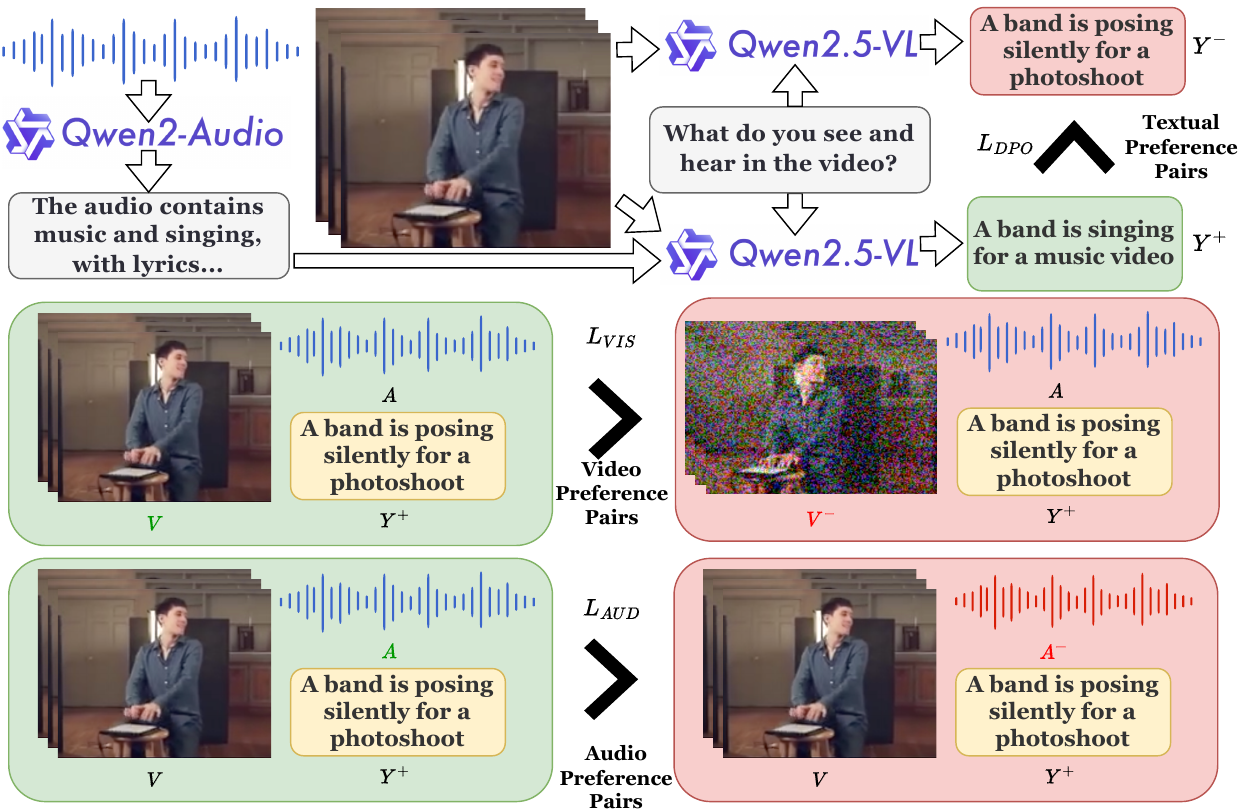}
  \caption{Overview of \modelname.}
  \label{fig:overview}
\end{figure}

To address the challenges of overly dominant textual priors and insufficient alignment between audio and visual modalities in omni-modal scenarios, we propose \modelname, a novel preference-based alignment framework that extends DPO \citep{rafailovDirectPreferenceOptimization2024} to omni-modal scenarios to effectively mitigate hallucinations across video, audio, and text modalities. We begin by constructing a new omni-modal preference-alignment dataset, \modelname-10k, to train our model. Each sample in the dataset consists of a question along with paired textual preferences and audio/video modality preferences, as illustrated in Figure \ref{fig:overview}. Based on the MSRVTT dataset \citep{xuMSRVTTLargeVideo2016}, we select video samples that contain audio. For generating positive textual preference samples, we first feed the audio into Qwen2-Audio \citep{chu2024qwen2audiotechnicalreport} to extract a description of its content. This audio-derived text is then combined with the video and passed to Qwen2.5-VL \citep{bai2025qwen25vltechnicalreport} to generate an answer that reflects understanding of the video's audio. For the negative samples, we input the same video without its audio into Qwen2.5-VL, producing an answer that lacks comprehension of the audio cues. These textual preference pairs specifically target the issue of misalignment between audio and visual modalities. Additionally, to address the problem of overly dominant textual priors, we introduce noisy variants of video and audio inputs to create modality preference pairs. These are used to encourage the model to attend more carefully to visual and auditory signals rather than relying solely on textual inputs. Based on this \modelname-10k, \modelname extends the Direct Preference Optimization (DPO) paradigm to multi-modal settings by introducing conditional preference learning for each modality. In addition to the standard DPO objective, \modelname incorporates modality-aware losses that explicitly encourage the model to attend to visual and auditory evidence.

Our experimental results demonstrate that applying \modelname to both Qwen2.5-Omni \citep{xu2025qwen25omnitechnicalreport} and MiniCPM-o-2.6 \citep{OpenBMB2025minicpm-o} leads to an average performance improvement of 3.48\% on the CMM benchmark \citep{leng2024curse} and 4.23\% on AVHBench \citep{sungbin2025avhbenchcrossmodalhallucinationbenchmark}. Notably, beyond effectively reducing omni-modal hallucinations, \modelname also enhances the model's reasoning and question-answering capabilities in single-modality scenarios. Our contributions can be summarized as:

• We propose \modelname, a direct preference optimization framework tailored for video-audio-text alignment, which addresses the challenge of hallucinations in omni-modal settings by introducing modality-specific conditional preference learning.

• We construct a 10k-sample omni-modal preference dataset \modelname-10k covering diverse real-world scenarios, designed to expose and counteract modality-specific and cross-modal hallucinations. To our best knowledge, \modelname and \modelname-10k are the first hallucination mitigation method and corresponding preference optimization dataset specifically designed for omni-modal scenarios.

• Extensive experiments on Qwen2.5-Omni and MiniCPM-o-2.6 demonstrate that \modelname significantly mitigates hallucinations in omni-modal settings while also enhancing the models' unimodal reasoning capabilities.

\section{Related Work}
\subsection{Omni-Modal Large Language Models}
Building on the success of large language models (LLMs) \citep{Brown2020, Llama2023, gpt4, GLM2024, chiang2023vicuna}, research in the multimodal domain has gained significant momentum. Omni-Modal Large Language Models utilize LLMs as their core language models, enhancing them with modality-specific tokenizers and encoders to map multimodal inputs into a shared representation space. The powerful language backbones then process these representations to generate outputs based on multimodal inputs and textual instructions \citep{Yin_2024}. Several studies \citep{bai2025qwen25vltechnicalreport, liu2023improvedllava,Xue2024,Liu2024b,Wang2024,lin2025uniworld} focus on enabling LLMs to handle image inputs, while others \citep{cheng2024videollama2advancingspatialtemporal,zhang2023videollamainstructiontunedaudiovisuallanguage,fu2024vitaopensourceinteractiveomni,yue2024coavtcognitioninspiredunifiedaudiovisualtext,chenVASTVisionAudioSubtitleTextOmniModality2023} aim to provide more comprehensive sensory capabilities to LLMs. These models are designed to process a variety of inputs, including but not limited to text, images, videos, and audio, while producing outputs in text or possibly audio formats as well \citep{OpenBMB2025minicpm-o,li2025baichuanomni15technicalreport,xu2025qwen25omnitechnicalreport,zhong2024lyraefficientspeechcentricframework,chen2025emovaempoweringlanguagemodels,su2023pandagptmodelinstructionfollow}. Some models even generate images and videos \citep{wu2024nextgptanytoanymultimodalllm,kondratyuk2024videopoetlargelanguagemodel,zhan2024anygptunifiedmultimodalllm} by incorporating modality-specific detokenizers or diffusion blocks. A significant challenge in this area is aligning the embeddings from different modalities into a unified representation space. While many works utilize a text-based or text-image joint space for this purpose, others experiment with anchoring the embeddings in the image space \citep{girdhar2023imagebindembeddingspacebind} or adopting an egalitarian approach to treat all modalities equally \citep{wangOmniBindLargescaleOmni2024}. Some studies have also proposed progressive alignment strategies  \citep{han2025onellmframeworkalignmodalities,liuOlaPushingFrontiers2025} to prevent forgetting and to better handle the interaction between modalities \citep{zhang2024extending,wang2024freebindfreelunchunified,wangOmniBindLargescaleOmni2024}.

\subsection{Hallucination in Omni-Modal Large Language Models}
While hallucinations in vision-language models have been extensively studied, relatively little research has focused on hallucinations in omni-modal models. Existing works on hallucinations in vision-language models point to several causes, such as modality gaps \citep{liang2022mindgapunderstandingmodality,liuSurveyHallucinationLarge2024}, language priors \citep{Huang2024}, and statistical biases \citep{agarwalCausalVQARevealing2020}. Mitigation efforts typically fall into two categories. The first approach focuses on additional training, including the creation of new datasets \citep{Liu2023, Gunasekar2023TextbooksAYN, chenPerturboLLaVAReducingMultimodal2025, Gunjal2024,sun2023aligninglargemultimodalmodels} or improving data cleaning techniques \citep{Yu2024b, Raunak2021CuriousCH, Shen2021IdentifyingUS}, as well as the introduction of new training objectives \citep{sarkar2025mitigatingobjecthallucinationmllms,chen2024alleviatinghallucinationslargevisionlanguage,zhouAligningModalitiesVision2024,zhaoHallucinationsEnhancingLVLMs2024}, most of which are addressing the issues of modal-inequality \citep{wangMDPOConditionalPreference2024,xieVDPOMitigatingHallucination2024,xiao2024seeingimageprioritizingvisual,Wu2024NoiseBoost,wu2025lanp} or even small modifications to the model structure \citep{yanTaskPreferenceOptimization2024,liao2025langbridge} or pipeline \citep{wuGenerateVerifyReducing2025}. The second approach targets the inference phase,
using methods such as contrast decoding, based on novel decoding strategies. The contrast term can be the logits resulting from distorted inputs \citep{ Chen2024HALC,lengMitigatingObjectHallucinations2023}, self-generated middle-term \citep{Chuang2024DoLa, woo2024dontmissforesttrees,huoSelfIntrospectiveDecodingAlleviating2025,liMitigatingHallucinationLarge2025a}, among others \citep{ Zhong2024, Kim2024, kim2024codecontrastingselfgenerateddescription, phan2024distillationcontrastivedecodingimproving, Park2024}. 
Another technique involves detecting and correcting hallucinations during generation \citep{Kuhn2023, Farquhar2024, Nikitin2024}, while some approaches manipulate the attention weights assigned to image inputs \citep{Zhu2024, Zhang2024, huo2024selfintrospectivedecodingalleviatinghallucinations,  An2024,liu2024payingattentionimagetrainingfree}.
Prompt-based strategies \citep{Qu2024LookCompareDecide, Wu2024LogicalClosedLoop, Xu2024ReReadingImprovesReasoning, Han2024, wang2024mitigatinghallucinationslargevisionlanguage} and
external tools \citep{Chern2023FacTool, Yin2023, Zhao2024} are also used to enhance the credibility of the model, along with inference time steering \citep{chen2024ictimageobjectcrossleveltrusted,liu2024reducinghallucinationsvisionlanguagemodels,jiang2025interpretingeditingvisionlanguagerepresentations,liHiddenLifeTokens2025}
and other strategies \citep{zou2024looktwiceanswermemoryspace,DBLP:conf/naacl/PanLLGDWKY25,DBLP:journals/corr/abs-2502-11598}. 
However, hallucinations in omni-modal LLMs remain an underexplored area, with a few notable exceptions. Research by  \cite{leng2024curse} and \cite{sungbin2025avhbenchcrossmodalhallucinationbenchmark} has identified modality imbalance as a contributing factor to hallucinations in OLLMs. Both works developed benchmarks to address this issue, but have not yet provided mitigation strategies.
\bibliographystyle{plain}

\section{Dataset Construction}
\label{data}
Multimodal hallucinations often arise because text priors dominate other modalities: even when a model receives visual and auditory inputs, it may ignore those signals in favor of familiar textual patterns.  Moreover, existing omni-modal training aligns each modality with text independently, overlooking the critical interaction between video and its native audio track.  As a result, when understanding hinges on subtle sound cues embedded in the footage, the model can confidently produce incorrect “hallucinated” answers.

To address these issues, we construct two types of preference pairs that explicitly reward genuine audio–visual reasoning:

\paragraph{Audio–video alignment preferences}  
To ensure our model truly leverages both audio and video, we first filter out any clip without an audio track, retaining only  $\mathcal{D}_0 = \bigl\{(V_i, A_i) \mid A_i \neq \emptyset\bigr\}$ from MSRVTT.  Each remaining sample is represented as \(X=\{V,A,T\}\), where \(V\) is the video frames, \(A\) the raw audio waveform, and \(T\) any accompanying text prompt or question.  We then call Qwen2-Audio, denoted  
\(\;t_a = \mathcal{F}_{\mathrm{audio}}(A)\), which produces a concise textual summary of speech, sound effects, or background music.  Feeding \(\{V,\,t_a\}\) into Qwen2.5-VL (denoted \(\mathcal{F}_{\mathrm{VL}}\)) yields a multimodal answer $Y^+ = \mathcal{F}_{\mathrm{VL}}(V,\,t_a),$ whereas masking out the audio text gives $
  Y^- = \mathcal{F}_{\mathrm{VL}}(V,\,\varnothing).$
The resulting pair $\{(X, Y^+),(X,Y^-)\}$ teaches the model that genuine audio cues in \(X\) should be preferred over video-only reasoning.

\paragraph{Modality-robustness preferences}  
Even with aligned audio–video pairs, large models tend to fall back on strong text priors.  To counteract this, we create two degraded versions of \(X\):
\[
  X_{V^-} = \{\,V^-,\,A,\,T\},\quad
  X_{A^-} = \{\,V,\,A^-,\,T\},
\]
where  
\begin{align}
  V^- &= V + \varepsilon_v,\quad \varepsilon_v\sim\mathcal{N}(0,\sigma_v^2 I),\\
  A^- &= A + \varepsilon_a,\quad \varepsilon_a\sim\mathcal{N}(0,\sigma_a^2 I).
\end{align}

\paragraph{Combined dataset}  
Putting both strategies together, our final dataset is  
\[
  \mathcal{D} \;=\;\bigl\{(X,Y^+),(X,Y^-)\bigr\}
  \;\cup\;\bigl\{(X,Y^+),(X_{V^-},Y^+)\bigr\}\;\cup\;\bigl\{(X,Y^+),(X_{A^-},Y^+)\bigr\}
,
\]
providing both audio–video alignment pairs and modality-robustness pairs to drive downstream preference optimization.

\begin{figure}[htbp]
  \centering
  \includegraphics[width=0.9\linewidth]{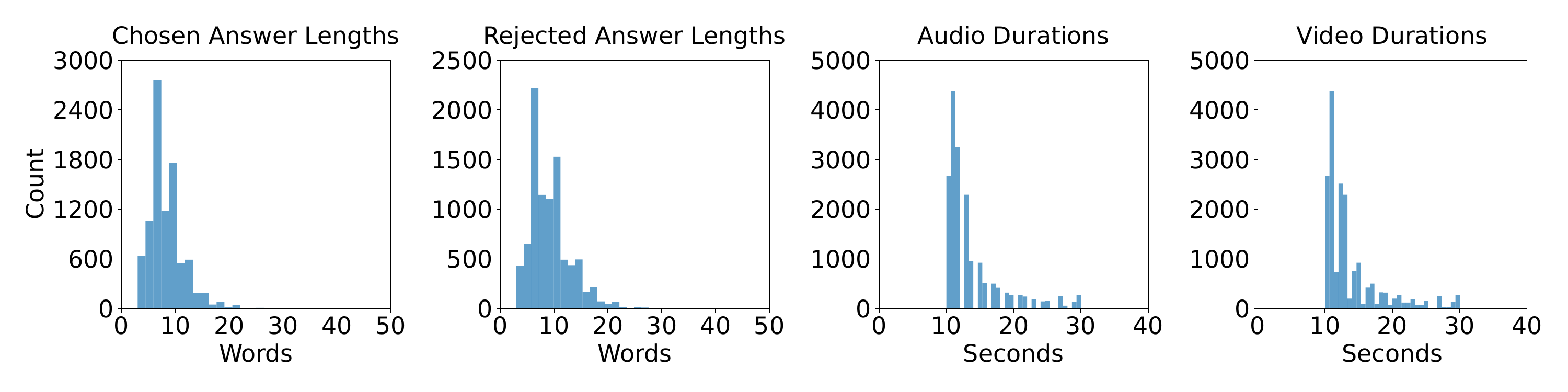}
  \caption{Key statistical information of our dataset.}
  \label{figure:hist_summary}
\end{figure}

In total, we collected 9141 pairs of samples on 1076 distinct videos. More elaborate information on durations of videos and audios and the length of the captions can be found in Figure \ref{figure:hist_summary}.

\section{Method}
\label{sec:method}
In this section, we detail the \modelname framework for omni-modal hallucination mitigation. We first formalize the problem and recap the standard DPO objective. We then introduce our conditional multi-modal preference optimization approach, describing how we generate modality-conditioned preference pairs and incorporate them into the training loss.
\subsection{Task Formulation}
We consider a general omni-modal input $X = \{V, A, T\}$, where $V$ is visual input (a sequence of video frames, which can degenerate to a single image), $A$ is auditory input (e.g. an audio track or spoken question), and $T$ is textual input (such as a text prompt or question). The OLLM is tasked with generating a textual output $Y$ (an answer or description). Hallucination occurs when $Y$ contains information not supported by $V$ or $A$. We assume access to a dataset of preference comparisons $\{(X, Y^+, Y^-)\}$, where for each input $X$ two candidate outputs are given: $Y^+$ (the preferred/correct output) and $Y^-$ (the less preferred, possibly hallucinated output). These could be obtained via human annotation or automated generation, plus human verification.

DPO provides a way to train the model $P_\theta(Y|X)$ to align with preferences without explicit reinforcement learning. Given a preference tuple $(X, Y^+, Y^-)$, the standard DPO objective encourages the model to increase the probability of $Y^+$ and decrease that of $Y^-$. The DPO loss is:
\begin{equation}
    L_{\text{DPO}}(\theta) = -\mathbb{E}_{(X, Y^+, Y^-)} \left[ \log \sigma \left( \beta \left( \frac{\log P_\theta(Y^+|X)}{\log P_{ref}(Y^+|X)} - \frac{\log P_\theta(Y^-|X)}{\log P_{ref}(Y^-|X)} \right) \right) \right],
\end{equation}

where $\sigma$ is the sigmoid function and $\beta$ is temperature for the preference difference. 
\subsection{\modelname}
As described in Section \ref{data}, in addition to the original human preference comparisons between output pairs $(Y^+, Y^-)$ given the full input $X$, we introduce new comparisons using modified inputs $X_{V^-}$ and $X_{A^-}$, which are constructed by masking visual or audio modalities, respectively. We denote the model’s output probabilities as $P_\theta(Y|X)$ when full input is given, and $P_\theta(Y|X_{V^-})$, $P_\theta(Y|X_{A^-})$ for the degraded cases.
\paragraph{Visual Preference Objective:}
 We create a comparison between the same output under full vs. degraded visual input. Ideally, the model should assign a higher likelihood to the correct output $Y^+$ when it has the full video $V$ than when it only has $V^-$. To enforce this, we treat $(X, Y^+)$ as the preferred scenario and $(X_{V^-}, Y^+)$ as the dispreferred scenario. We do not alter the output in this pair — only the input differs. The visual preference loss $L_{\text{vis}}$ can be written as:
 \begin{equation}
L_{\mathrm{vis}}(\theta)
=
-\,\mathbb{E}_{(X,X_{V^-},Y^+)}\!
\left[
\log\sigma\!\Bigl(
\beta\Bigl[
\log\frac{P_\theta(Y^+\mid V,A,T)}{P_{\mathrm{ref}}(Y^+\mid V,A,T)}
-
\log\frac{P_\theta(Y^+\mid V^-,A,T)}{P_{\mathrm{ref}}(Y^+\mid V^-,A,T)}
\Bigr]
\Bigr)
\right]
 \end{equation}
 This term pushes the model to increase $P_\theta(Y^+|V,A,T)$ relative to $P_\theta(Y^+|V^-,A,T)$. Intuitively, the model is penalized if it would be just as confident in $Y^+$ even with a missing/blurred video. The only way for the model to satisfy this objective is to genuinely utilize the visual input $V$ when available (since with $V$ it should produce higher confidence in the correct answer). Notably, we do not include $Y^-$ in this comparison; we are not directly contrasting $Y^+$ vs $Y^-$ here, but rather $Y^+$ with vs. without vision.
 \paragraph{Auditory Preference Objective:} Analogously, we define an audio preference loss $L_{\text{aud}}$. We compare the model’s confidence in $Y^+$ with full input vs. with muted audio. The audio preference loss is:
 \begin{equation}
L_{\mathrm{aud}}(\theta)
=
-\,\mathbb{E}_{(X,X_{A^-},Y^+)}\!
\left[
\log\sigma\!\Bigl(
\beta\Bigl[
\log\frac{P_\theta(Y^+\mid V,A,T)}{P_{\mathrm{ref}}(Y^+\mid V,A,T)}
-
\log\frac{P_\theta(Y^+\mid V,A^-,T)}{P_{\mathrm{ref}}(Y^+\mid V,A^-,T)}
\Bigr]
\Bigr)
\right]
 \end{equation}
 This term teaches the model to rely on auditory information $A$ when it is present. For example, if the question $T$ asks “Is the person in the video speaking?” and the correct answer $Y^+$ is “Yes, you can hear them speak”, then with $A$ silenced the model should be much less certain, and $L_{\text{aud}}$ will enforce that difference in probability. Without such training, a model might answer “Yes” purely from prior knowledge that people in videos often speak, which would be a hallucination if the audio is silent; \modelname prevents this by explicitly penalizing confidence when the evidence (audio) is absent. It’s important to note that the degraded-input comparisons always involve the preferred output $Y^+$. We assume $Y^+$ is the ground-truth or a human-verified correct answer that is consistent with the full input. By training the model on $(X_{V^-}, Y^+)$ and $(X_{A^-}, Y^+)$, we are not asserting that $Y^+$ is wrong for the degraded input—rather, we are asserting that with less information, the model should be less sure. In practice, if $Y^+$ truly depends on the missing information, $Y^+$ might no longer even be a correct answer when the modality is removed (e.g., if the video is removed, the question might be unanswerable). The model doesn’t explicitly know this, but the optimization will implicitly encourage the model to assign lower probability in those cases, which aligns with the idea that it shouldn’t output $Y^+$ confidently if it can’t see/hear the necessary cues.

 \paragraph{Combining Objectives:} The full \modelname loss combines the standard preference loss with the new modality-specific terms, where $\lambda_V$ and $\lambda_A$ are hyperparameters:
 \begin{equation}
     L_{\text{OMNI}}(\theta) = L_{\text{DPO}}(\theta) + \lambda_V L_{\text{vis}}(\theta) + \lambda_AL_{\text{aud}}(\theta).
 \end{equation}
\section{Experiments}
\subsection{Experiments Setup}
\label{subsec:setup}
\paragraph{Datasets and Metrics.}

\noindent\textbf{AVHBench} \cite{sungbin2025avhbenchcrossmodalhallucinationbenchmark} is specifically designed to assess the perception, reasoning, and hallucination robustness of OLLMs. It comprises four evaluation subsets: Audio-Visual Matching, Audio-Visual Captioning, Audio-Driven Video Hallucination, and Video-Driven Audio Hallucination. In our evaluation, we focus on the two hallucination-focused subsets, which specifically measure a model’s tendency to hallucinate visual content based solely on audio priors, or vice versa. To quantify performance in hallucination mitigation, the benchmark employs standard classification metrics: Accuracy, Precision, Recall, and F1 Score.

\noindent\textbf{CMM} (Curse of Multi-Modalities) \citep{leng2024curse}is a benchmark specifically designed to evaluate hallucination behavior in OLLMs across text, vision, and audio modalities. It focuses on measuring a model’s tendency to generate modality-inconsistent outputs by presenting it with inputs where one or more modalities contradict the others. Each sample is paired with two probing questions—one targeting an existing object or event (ground-truth answer "yes") and one targeting a non-existent one (ground-truth "no")—covering both visual and auditory modalities. CMM uses two core metrics: Perception Accuracy (PA) and Hallucination Resistance (HR), defined as follows:
\begin{equation}
\text{PA} = \frac{\text{\# correctly predicted “yes”}}{\text{\# ground-truth “yes”}}, \quad
\text{HR} = \frac{\text{\# correctly predicted “no”}}{\text{\# ground-truth “no”}},
\end{equation}
where PA measures the model’s ability to correctly detect real elements, while HR quantifies its robustness in rejecting non-existent ones.

\paragraph{Baselines and Models.}
We conduct experiments using two recently released and widely adopted OLLMs: Qwen2.5-Omni and MiniCPM-o-2.6. As baselines, we include both training free methods such as VCD and ICT, as well as a training based method, DPO (text only), which performs preference optimization solely on textual responses.

\paragraph{Implention Details.}
For both models, we adopt full-parameter fine-tuning using fp16 precision. The learning rate scheduler is set to cosine, with a warmup ratio of 0.1. The DPO loss coefficient ($\beta$) is fixed at 0.1 throughout all training runs. For Qwen2.5-Omni, we use a learning rate of 1e-6, while for MiniCPM-o-2.6, the learning rate is set to 1e-5. We set $\lambda_V = \lambda_A = 1$. All experiments were conducted
on a system equipped with 8 × H100 GPUs
\subsection{Main Results}
\begin{table}[b]
  \caption{AVHBench results for Qwen2.5-Omni and MiniCPM-o-2.6 on the two subsets—Audio-driven Video Hallucination and Video-driven Audio Hallucination. Metrics include Accuracy (Acc.), Precision (Prec.), Recall (Rec.), F1-score, and the proportion of “Yes” responses (Yes \%)}. 
  \centering
  \setlength{\tabcolsep}{1.0mm}
  \resizebox{1\linewidth}{!}{\begin{tabular}{lcccccccccc}
    \toprule
    \multirow{2}{*}[-0.4em]{Model} & \multicolumn{5}{c}{\textbf{Audio-driven Video Hallucination}} & \multicolumn{5}{c}{\textbf{Video-driven Audio Hallucination}} \\
    \cmidrule(r{2mm}l{2mm}){2-6} \cmidrule(r{2mm}l{2mm}){7-11}
    
    &Acc. ($\uparrow$)&Prec. ($\uparrow$)&Rec. ($\uparrow$)&F1 ($\uparrow$)&Yes (\%)&Acc. ($\uparrow$)&Prec. ($\uparrow$)&Rec. ($\uparrow$)&F1 ($\uparrow$)&Yes (\%)\\
    \hline
    \rowcolor{gray!20} \multicolumn{11}{c}{Qwen2.5-Omni} \\
    \hline
Qwen2.5-Omni & 74.12 & 68.72 & \textbf{88.56} & 77.38 & 64.44 & 67.60 & 60.84 & \textbf{98.78} & 75.30 & 81.18 \\
+ VCD        & 77.40 & 75.94 & 80.20 & 77.52 & 60.02 & 68.11 & 61.40 & 97.52 & 77.38 & 79.90 \\
+ ICT        & 76.70 & 73.12 & 84.45 & 78.58 & 61.28 & 68.80 & 62.10 & 96.50 & 77.80 & 78.95 \\
+ DPO        & 71.74 & 67.73 & 83.04 & 74.61 & 58.19 & 68.70 & 62.41 & 94.04 & 75.03 & 71.72 \\
\rowcolor{green!20} + \modelname & \textbf{84.42} & \textbf{88.87} & 78.70 & \textbf{83.47} & 44.28 & \textbf{77.51} & \textbf{70.40} & 94.93 & \textbf{80.85} & 67.42 \\
    \hline
    \rowcolor{gray!20} \multicolumn{11}{c}{MiniCPM-o-2.6} \\
    \hline
MiniCPM-o-2.6   & 74.65 & 72.78 & 78.75 & 75.71 & 53.79 
                & 72.80 & 70.42 & 78.63 & 74.41 & 55.55 \\
+ VCD           & 73.97 & 72.94 & 76.20 & 74.55 & 53.21 
                & 73.96 & 72.15 & 78.05 & 75.04 & 55.05 \\
+ ICT           & 75.92 & 74.83 & 78.12 & 76.46 & 53.32 
                & 72.25 & 69.88 & 78.20 & 73.92 & 55.18 \\
+ DPO           & 76.07 & 74.66 & 78.92 & 76.76 & 52.55 
                & 74.73 & 72.81 & 78.93 & 75.81 & 54.54 \\
\rowcolor{green!20} +\modelname     & \textbf{76.85} & \textbf{74.72} & \textbf{81.16} & \textbf{77.81} & 54.31 
                & \textbf{76.68} & \textbf{74.70} & \textbf{80.70} & \textbf{77.58} & 54.02 \\

  \bottomrule
  \end{tabular}
  }
  \centering
  \label{tab:crossmodal_halluci}
\end{table}
\begin{table}
\caption{ Performance of Qwen2.5-Omni and MiniCPM-o-2.6 on the CMM benchmark after applying different hallucination-mitigation strategies.
The benchmark examines two broad error sources—Spurious Inter-modality Correlations and Over-reliance on Unimodal Priors—broken down into six sub-domains: VL (Visual–Language), AL (Audio–Language), VAL (Visual–Audio–Language), Visual Dom (Visual-dominance), Audio Dom (Audio-dominance), and Lang Dom (Language-dominance).}

\centering
\renewcommand{\arraystretch}{1.2}
\begin{adjustbox}{width=\linewidth}
\setlength{\tabcolsep}{1.2mm}
\small
\begin{tabular}{l|cc|cc|cc|cc|cc|cc|cc}
\noalign{\hrule height 1.5pt}
\multicolumn{1}{c|}{\multirow{3}{*}{Model}} & \multicolumn{6}{c|}{Spurious Inter-modality Correlation}  & \multicolumn{6}{c|}{Uni-modality Overreliance} & \multicolumn{2}{c}{Overall} \\ 
\cline{2-15}
& \multicolumn{2}{c|}{VL} & \multicolumn{2}{c|}{AL} & \multicolumn{2}{c|}{VAL} & \multicolumn{2}{c|}{Visual Dom} & \multicolumn{2}{c|}{Audio Dom} & \multicolumn{2}{c|}{Lang Dom} & \multirow{2}{*}{pa~$\uparrow$}        & \multirow{2}{*}{hr~$\uparrow$}   \\
& pa~$\uparrow$ & hr~$\uparrow$ & pa~$\uparrow$ & hr~$\uparrow$ & pa~$\uparrow$ & hr~$\uparrow$ & pa~$\uparrow$ & hr~$\uparrow$ & pa~$\uparrow$ & hr~$\uparrow$ & pa~$\uparrow$ & hr~$\uparrow$ & &\\
\hline
\rowcolor{gray!20} \multicolumn{15}{c}{Qwen2.5-Omni} \\
\hline

\multicolumn{1}{l|}{Qwen2.5-Omni}         
& 92.0 & 86.5 
& \textbf{92.0} & 78.0 
& 93.0 & 90.5 
& 95.0 & 56.5 
& 92.5 & 39.5 
& 85.5 & 74.0 
& 91.7 & 70.2 \\

\multicolumn{1}{l|}{+VCD}       
&93.5  & 89.0 
&90.0  & 77.5 
&93.5  & 90.0 
&94.5  & 56.5  
&93.0  & 40.0 
&87.5  & 75.0 
&92.0  & 71.3 \\

\multicolumn{1}{l|}{+ICT}       
&\textbf{94.5}  &\textbf{90.0 } 
&89.0  &78.0  
&94.0  &91.5  
&93.5  &57.0  
&93.5  &39.5  
&\textbf{88.0}  &75.0  
&92.1  &71.8  \\

\multicolumn{1}{l|}{+DPO}       
& 92.5 & 87.0 
& 92.0 & 79.5 
& 95.0 & 93.5 
& \textbf{95.5} & \textbf{66.0} 
& 94.5 & 37.5 
& 87.0 & \textbf{76.0 }
& 92.8 & 73.3 \\

\rowcolor{green!20}\multicolumn{1}{l|}{+\modelname}       
& 94.0 & 89.5 
& 91.5 & \textbf{83.5} 
& \textbf{95.5} & \textbf{94.5} 
& \textbf{95.5} & 63.5 
& \textbf{95.0} & \textbf{41.5} 
& \textbf{88.0} & 75.5 
& \textbf{93.3} & \textbf{74.7} \\
\hline
\rowcolor{gray!20} \multicolumn{15}{c}{MiniCPM-o-2.6} \\ 
\hline
MiniCPM-o-2.6              & 85.0 & 91.0 & 95.0 & 53.0 & 92.0 & 76.5 & 91.0 & 56.5 & 89.0 & 34.5 & 76.5 & 72.0 & 88.1 & 63.9 \\
+VCD                       & 88.0   & 91.5   & 95.0   & 52.0   & 91.0   & 76.5   &  89.5  & 55.0   & 89.0   & 33.0   & 78.5   & 74.5   & 88.5   & 63.8   \\
+ICT                       & 89.0   & 93.5   & 94.5   & 54.5   & 93.0   & 76.5   & 88.5   & 56.0   & 90.5   & \textbf{35.0}   & 78.0   & 75.5   & 88.9   & 65.1   \\
+DPO                       & 87.0 & 93.0 & 95.5 & 57.5 & \textbf{94.5} & 78.5 & 90.5 & 58.5 & \textbf{91.5} & 32.5 & \textbf{81.5} & 77.0 & 90.1 & 66.2 \\
\rowcolor{green!20}+\modelname                & \textbf{90.0} & \textbf{94.0} & \textbf{96.5} & \textbf{59.5} & 93.0 & \textbf{81.5} & \textbf{94.0} & \textbf{63.5} & \textbf{91.5} & 32.5 & 81.0 & \textbf{82.0} & \textbf{91.0} & \textbf{68.8} \\

\noalign{\hrule height 1.5pt}
\end{tabular}
\end{adjustbox}
\vspace{-1mm}
\vspace{-3mm}
\label{tab:CMM}

\end{table}
\paragraph{AVHBench} Table \ref{tab:crossmodal_halluci} presents the results of Qwen2.5-Omni and MiniCPM-o-2.6 on the two domains of AVHBench: Audio-Driven Video Hallucination and Video-Driven Audio Hallucination. From the results, we draw the following conclusions: \textbf{1)} Applying \modelname leads to significant improvements in F1-score, with average gains of 5.82\% for Qwen2.5-Omni and 2.64\% for MiniCPM-o-2.6. This demonstrates that \modelname, by leveraging both textual preference optimization and omni-modal preference alignment, effectively mitigates hallucination in multimodal settings. \textbf{2)} Existing methods such as VCD and ICT, which were originally proposed for vision-language hallucination mitigation, provide only marginal improvements—and in some cases even degrade performance. While these methods reduce the influence of biased language priors and encourage visual grounding, they fail to model the subtle interactions between audio and visual modalities. Moreover, intervention-based approaches like ICT rely on activation shifts observed in visual tasks, which do not generalize well to omnimodal contexts. \textbf{3)} For Qwen2.5-Omni, the base model exhibits a strong bias toward answering "yes," often relying on a single modality without properly integrating cross-modal information. This results in hallucinations when critical multimodal cues are absent. \modelname addresses this by constructing multimodal preference pairs that teach the model to be cautious in its affirmations. As a result, the "yes" response rate drops by 16.96\%, indicating improved resistance to hallucination in omni-modal scenarios. \textbf{4)} While text-only DPO helps the model better understand fine-grained audio-video relationships, it fails to sufficiently shift the model’s attention toward non-textual modalities. The dominance of the language modality remains unbalanced, limiting its effectiveness and resulting in no significant performance gains.
\paragraph{CMM} Table~\ref{tab:CMM} presents the performance of Qwen2.5-Omni and MiniCPM-o-2.6 on the CMM benchmark, which evaluates hallucination robustness under two major domains: Spurious Inter-modality Correlation and Uni-modality Overreliance. Based on the results, we draw the following conclusions: \textbf{1)} Applying \modelname consistently improves both PA and HR across all sub-domains. For Qwen2.5-Omni, \modelname achieves an average improvement of +1.6\% (pa) and +4.5\% (hr). Similarly, MiniCPM-o-2.6 sees gains of +1.9\% (pa) and +4.9\% (hr). Notably, the larger improvement in hr suggests that \modelname is particularly effective at reducing false affirmations of non-existent multimodal information, which aligns with its design: by training on modality-aware preference pairs, the model learns to suppress overconfident hallucinations when visual or auditory evidence is insufficient. \textbf{2)} Methods like VCD and ICT, which focus on vision-language grounding, improve performance in the VL domain, but this comes at the cost of decreased performance in other areas. Specifically, these methods often degrade performance in the Visual Dominance domain, suggesting an over-correction toward visual features that suppresses necessary cross-modal balance. Moreover, since neither VCD nor ICT explicitly models audio-visual interactions, they fail to improve the VAL (video-audio-language) sub-domain. This highlights the limitation of vision-centric interventions when applied to omni-modal contexts. \textbf{3)} While standard DPO slightly improves VAL by learning from textual preference pairs, it brings minimal improvements—or even regressions—on other domains. For example, DPO underperforms in Audio Dominance and Visual Dominance, indicating that without modality-specific conditioning, models still over-rely on dominant priors or shortcuts. In contrast, \modelname improves all sub-domains by explicitly encouraging the model to distinguish between complete and degraded modality inputs.
 \section{Analysis}
\subsection{Ablation Study}
To investigate how preference optimization for different modalities affects a model’s resistance to hallucination, we conduct an ablation study on Qwen2.5-Omni. The results are presented in Table \ref{tab:ablation}. From the table, we observe that applying either video-only or audio-only preference optimization individually yields positive effects on the model’s performance. When both modalities are optimized simultaneously, the model's robustness against hallucination is further enhanced. Interestingly, on the CMM benchmark, we notice a modality interference effect: optimizing preferences for only one modality can negatively impact the performance related to the other modality. In contrast, jointly optimizing both audio and video preferences leads to consistent and overall improvements across modalities.
\begin{table}[b]
\vspace{-5mm}
\caption{Ablation Study. “+ Audio” denotes applying only the audio loss $L_{\text{aud}}$, while “+ Video” uses only the visual loss $L_{\text{vis}}$. For the CMM dataset, results are reported on the Spurious Inter-modality Correlation subsets: VL, AL, and VAL. For AVHBench, “A-V Hal.” refers to the Audio-driven Video Hallucination setting, and “V-A Hal.” to the Video-driven Audio Hallucination setting.}

\centering
\begin{adjustbox}{width=0.8\linewidth}
\setlength{\tabcolsep}{1.2mm}
\footnotesize
\begin{tabular}{l|cccccc|cccc}
\noalign{\hrule height 1.5pt}
\multicolumn{1}{c|}{\multirow{3}{*}{Model}} & \multicolumn{6}{c|}{CMM} & \multicolumn{4}{c}{AVHBench} \\
& \multicolumn{2}{c|}{VL} & \multicolumn{2}{c|}{AL} & \multicolumn{2}{c|}{VAL} & \multicolumn{2}{c|}{A-V Hal.} & \multicolumn{2}{c}{V-A Hal.} \\
& pa~$\uparrow$ & hr~$\uparrow$ & pa~$\uparrow$ & hr~$\uparrow$ & pa~$\uparrow$ & hr~$\uparrow$ & Acc.~$\uparrow$ & F1~$\uparrow$ & Acc.~$\uparrow$ & F1~$\uparrow$ \\
\hline
\multicolumn{1}{l|}{Qwen2.5-Omni}
& 92.0 & 86.5
& 92.0 & 78.0
& 93.0 & 90.5
& 74.1 & 77.4
& 67.6 & 75.3 \\

\multicolumn{1}{l|}{+ Audio}
& 93.5 & 82.5
& \textbf{93.0} & 82.5
& 92.5 & 93.5
& 80.9 & 81.1
& 76.7 & 80.4 \\

\multicolumn{1}{l|}{+ Video}
& \textbf{95.5} & 88.5
& 90.5 & 63.0
& 94.0 & 66.0
& 82.4 & 81.4
& 66.3 & 74.6 \\

\multicolumn{1}{l|}{+ \modelname}
& 94.0 & \textbf{89.5}
& 91.5 & \textbf{83.5}
& \textbf{95.5} & \textbf{94.5}
& \textbf{84.4} & \textbf{83.5}
& \textbf{77.5} & \textbf{80.9} \\
\hline

\noalign{\hrule height 1.5pt}
\end{tabular}
\end{adjustbox}
\vspace{-1mm}
\vspace{-3mm}
\label{tab:ablation}

\end{table}

\subsection{Experiments on Non-hallucination Benchmarks}
\begin{figure}[htbp]
  \centering
  \includegraphics[width=0.9\linewidth]{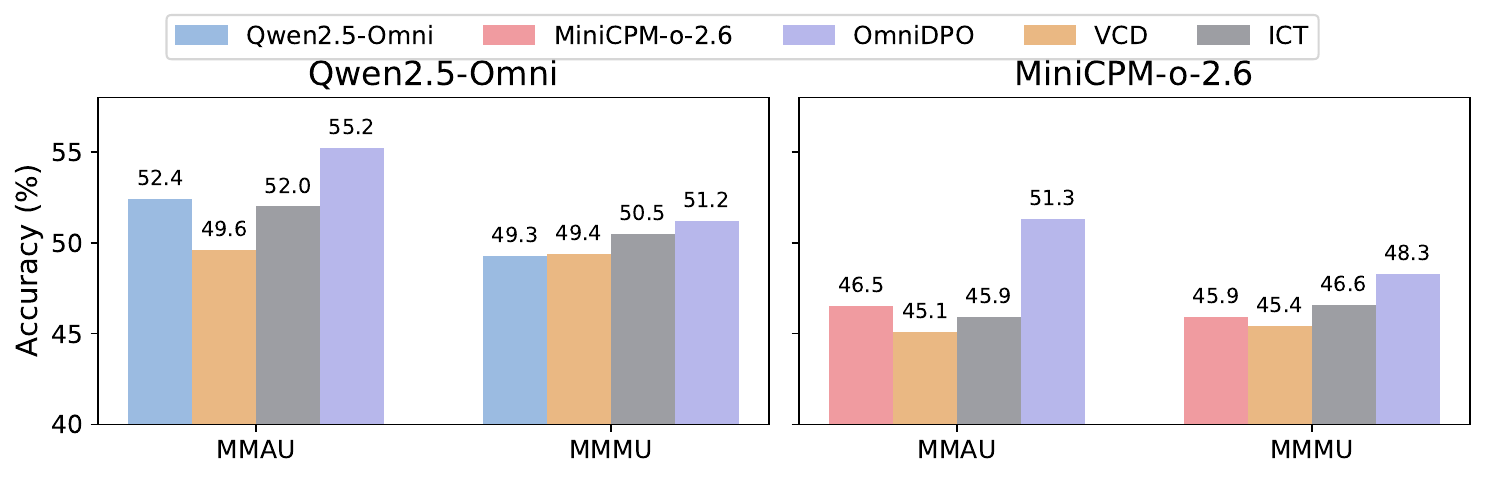}
  \vspace{-5mm}
  \caption{Performance comparison on reasoning benchmarks (MMAU \citep{sakshi2024mmaumassivemultitaskaudio} and MMMU \citep{yue2023mmmu}) after applying different alignment methods. All evaluations are performed in a zero-shot setting. \modelname leads to consistent improvements across both benchmarks, indicating enhanced multimodal reasoning, while VCD and ICT show limited or modality-specific gains.}
  \label{figure:mmu}
  \vspace{-5mm}
\end{figure}
To investigate whether applying \modelname affects a model’s general reasoning capabilities, we evaluate performance on two widely used benchmarks: MMAU (Massive Multi-Task Audio Understanding) \citep{sakshi2024mmaumassivemultitaskaudio} and MMMU (Massive Multi-discipline Multimodal Understanding) \citep{yue2023mmmu}. We also compare results with VCD and ICT. The outcomes are shown in Figure \ref{figure:mmu}. Experimental results show that after applying \modelname, Qwen2.5-Omni achieves an average performance gain of 2.4\%, while MiniCPM-o-2.6 improves by 3.6\%. This indicates that \modelname not only mitigates hallucinations but also enhances the model’s reasoning and understanding capabilities by encouraging better attention to audio and visual inputs through multimodal preference optimization. In contrast, VCD reduces performance on reasoning tasks, likely because it removes language priors, including those beneficial for reasoning. ICT, while improving performance on MMMU by reinforcing visual grounding, does not account for audio modality and thus fails to improve the model’s understanding of auditory information.
\subsection{Case Study and Error Analysis}
\begin{figure}[htbp]
  \centering
  \vspace{-3mm}
  \includegraphics[width=0.9\linewidth]{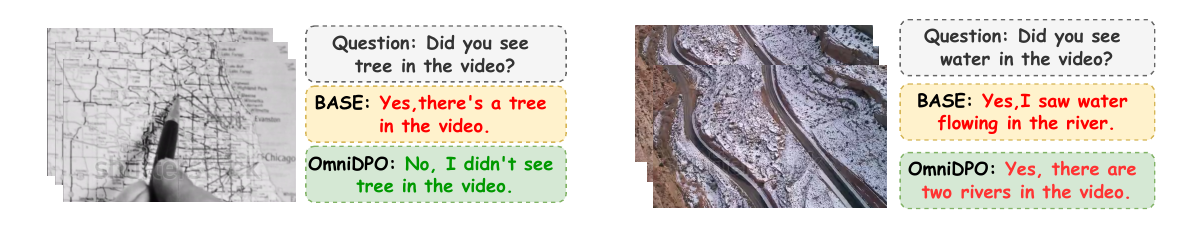}
  \caption{Case Study and Error Analysis of \modelname.}
  \label{figure:casestudy}
  \vspace{-3mm}
  
\end{figure}
In Figure \ref{figure:casestudy}, we present a case study from the CMM benchmark to evaluate the performance of our model, \modelname. As illustrated on the left side of the figure, while the base model, Qwen2.5-Omni, failed to distinguish between maps with trees, our \modelname completed the task. In this instance, \modelname overcame the tendency to answer "Yes" due to language priors and instead derived the correct response from the visual information.

On the right side of the figure, however, both the base model and \modelname produced incorrect answers. We hypothesize that this is due to limitations in the base model’s visual understanding. Although \modelname mitigates the issue of over-reliance on textual modality, it does not fundamentally enhance the visual capabilities inherited from the base model. Consequently, when faced with ambiguous visual content, such as in the right part of Figure \ref{figure:casestudy}, where roads closely resemble rivers, the model still struggles to provide an accurate response.
\section{Conclusion and Limitations}
\label{sec:conclusion}
We presented \modelname, the first dedicated framework for mitigating omni-modal hallucinations in OLLMs. By extending DPO with modality-specific objectives, \modelname effectively compels models to ground their outputs in both visual and auditory inputs, rather than over-relying on textual priors or spurious correlations. To support training, we constructed OmniDPO-10k, the first preference-alignment dataset specifically designed to expose and counteract hallucinations arising from misaligned or over-relied modalities (text, video, audio). Each sample includes paired modality-based preferences, enabling fine-grained supervision over model behavior. Through experiments on cutting-edge benchmarks (CMM and AVHBench) and models of different scales, we demonstrated that \modelname substantially reduces hallucinations across video, audio, and text scenarios, outperforming existing alignment methods.

\paragraph{Limitations} Due to limitations in the base models, our work currently supports only textual, visual, and audio modalities. Future research may explore the incorporation of more diverse input forms. Although our methods involve training, the computational cost is moderate rather than prohibitive; nonetheless, the development of training-free approaches could offer valuable alternatives. 

\bibliography{references}
\clearpage
\thispagestyle{empty}   
\setcounter{page}{1} 
\begin{center}
  {\large\bfseries OMNIDPO: A Preference Optimization Framework to
Address Omni-Modal Hallucination}\\[0.5em]
\end{center}

\section*{Appendix A: Baseline Methods
}
\paragraph{VCD} VCD \citep{lengMitigatingObjectHallucinations2023} is a training-free method that mitigates hallucinations by contrasting the logits derived by the original image and the distorted image, which can reduce over-reliance on statistical bias and unimodal priors. 
\paragraph{ICT} ICT \citep{chen2024ictimageobjectcrossleveltrusted} utilizes a set of pre-calculated intervention vectors during inference time. These vectors capture the directions in activation space that motivate the model to pay more attention to both overall and fine-grained visual information. Through applying these intervention vectors on specific attention heads, ICT mitigates harmful language priors, thereby alleviating hallucinations.
\section*{Appendix B: Codes and Datasets
}
For the official implemention of our \modelname as well as the \modelname-10k, please refer to \url{https://anonymous.4open.science/r/OmniDPO-7E3C}
\section*{Appendix C: More Information About \modelname-10k
}
In Table \ref{tab:appendix-examples}, we provide some examples of chosen outputs and rejected outputs in our \modelname-10k. All the samples share the same question: \texttt{<video><audio> What do you see and hear?}

\begin{figure}[htbp]
  \centering
  \includegraphics[width=0.9\linewidth]{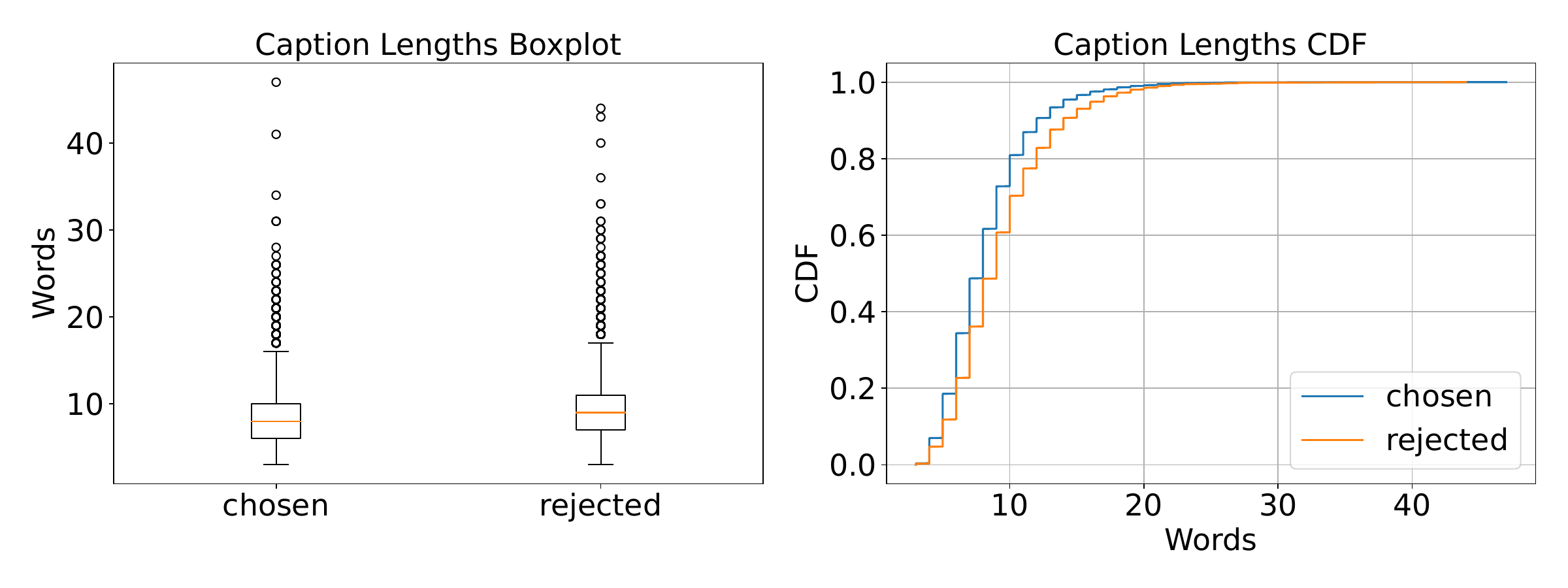}
  \caption{Key statistical information of our dataset.}
  \label{figure:combined_captions}
\end{figure}

\begin{table}[htbp]
  \centering
  \small
  \caption{Example Model Outputs (Chosen vs.\ Rejected)}
  \label{tab:appendix-examples}
  \begin{tabular}{p{0.45\textwidth} p{0.45\textwidth}}
    \toprule
    \textbf{Chosen Output} & \textbf{Rejected Output} \\
    \midrule
    “a person is explaining something” & “a person is singing something” \\[0.5em]
    “the dogs on Paw Patrol are listening to a mission” & “the dogs on Paw Patrol are watching a mission” \\[0.5em]
    “a man narrates a WF match between two men in a wrestling ring” & “a man gestures enthusiastically while observing a WF match between two men in a wrestling ring” \\[0.5em]
    “a band is singing for a music video” & “a band is posing silently for a photoshoot” \\
    \bottomrule
  \end{tabular}
\end{table}
In Figure \ref{figure:combined_captions}, we provide more statistical information about our dataset.

\end{document}